\documentclass{article} % For LaTeX2e
\PassOptionsToPackage{options}{natbib}
\usepackage[nonatbib,final]{neurips_2025}
\usepackage{hyperref}
\usepackage{url}
\usepackage{graphicx}
\usepackage{amsmath}
\usepackage{amsfonts}
\usepackage{subcaption}
\usepackage{biblatex}
\usepackage{enumitem}

\usepackage{xcolor}
\begin{document}

\title{Are We Really Learning the Score Function? Reinterpreting Diffusion Models Through \\ Wasserstein Gradient Flow Matching}

% Authors must not appear in the submitted version. They should be hidden
% as long as the \iclrfinalcopy macro remains commented out below.
% Non-anonymous submissions will be rejected without review.

\author{
An B. Vuong$^{1,\dagger}$ \quad Michael T.~McCann$^{2}$ \quad Javier E. ~Santos$^{2,\dagger}$ \quad Yen Ting Lin$^{2,\dagger,\ast}$\\
$^1$Oregon State University \quad $^2$Los Alamos National Laboratory \\
$^\dagger$Applied Machine Learning Summer School \quad $^\ast$Corresponding author\\
\quad vuonga2@oregonstate.edu, \{jesantos,mccann,yentingl\}@lanl.gov\\
}
\maketitle

\begin{abstract}
Diffusion models are commonly interpreted as learning the \emph{score function}, i.e., the gradient of the log-density of noisy data. However, this assumption implies that the target of learning is a conservative vector field, which is not enforced by the neural network architectures used in practice.
We present numerical evidence that trained diffusion networks violate both integral and differential constraints required of true score functions, demonstrating that the learned vector fields are not conservative. Despite this, the models perform remarkably well as generative mechanisms.  To explain this apparent paradox, we advocate a new theoretical perspective: diffusion training is better understood as \emph{flow matching} to the velocity field of a Wasserstein Gradient Flow (WGF), rather than as score learning for a reverse-time stochastic differential equation. 
Under this view, the ``probability flow'' arises naturally from the WGF framework, eliminating the need to invoke reverse-time SDE theory and clarifying why generative sampling remains successful even when the neural vector field is not a true score. We further show that non-conservative errors from neural approximation do not necessarily harm density transport.  Our results advocate for adopting the WGF perspective as a principled, elegant, and theoretically grounded framework for understanding diffusion generative models.
\end{abstract}

\section{Background}
%\subsection{Diffusion Models}
Diffusion models are typically described as follows: 
Given $D$-dimensional samples $x \in \mathbb{R}^D$ drawn from a data distribution $\mu_0$, one defines a forward It\^o process that gradually corrupts $x$ into noise.
Throughout this paper, we use the continuous-time Ornstein--Uhlenbeck (OU) process for concreteness:\footnote{
\cite{santos_understanding_2023} established the equivalence of the OU process with the discrete-time Denoising Diffusion Probabilistic Model \cite{hoDenoisingDiffusionProbabilistic2020} and the score-based formulation \cite{songScoreBasedGenerativeModeling2021a}. 
This setup is often called ``variance-preserving'' (VP), though this term is misleading: for each sample, the variance is not constant over time (which, in most scientific contexts, is the definition of ``preserving''), but grows as $\sqrt{1-e^{-2t}}$. Our analysis extends naturally to the standard Brownian motion process $\text{d}X_t = \text{d}W_t$, commonly termed ``variance-exploding'' (VE).} 
\begin{equation}
    \text{d} X_t = - X_t \, \text{d}t + \sqrt{2}\, \text{d}W_t, 
    \qquad X_{0} = x \sim \mu_0, \label{eq:OUprocess}
\end{equation}
where each component of $W_t$ is a standard Wiener process. The process \eqref{eq:OUprocess} converges to a limiting distribution $\mu_\infty$ as $t \to \infty$, which is an isotropic Gaussian in $\mathbb{R}^D$. Because of the choice of diagonal matrices in the drift and diffusion terms, each component of $X_t$ follows the well-studied one-dimensional OU process.

Equivalently, the forward dynamics can be described in terms of densities. The transition kernel\footnote{Since the OU process decomposes into $D$ independent one-dimensional processes, the density factorizes across coordinates: $\rho(\xi,t\mid \zeta,s) = \prod_{i=1}^D \rho_i(\xi_i, t\mid \zeta_i,s)$} $\rho(\xi, t \vert \zeta, s)$ satisfies the Fokker--Planck Equation (FPE):
\begin{equation}
    \partial_t \rho(\xi,t \vert \zeta,s) 
    = \nabla_\xi \!\left[\xi \, \rho(\xi,t \vert \zeta,s)\right] 
      + \nabla_\xi^2 \rho(\xi,t \vert \zeta,s),
    \label{eq:FP}
\end{equation}
with the initial condition $\rho(\xi,0) = \delta(\xi - x)$ for each of the drawn samples $x \sim \mu_0$, where $\delta(\cdot)$ denotes the Dirac delta distribution.

The modern understanding of diffusion models is grounded in Anderson's reverse-time theory \cite{anderson_reverse-time_1982}, which guarantees the existence of a reverse-time It\^o process that transforms samples from the simple distribution $\mu_\infty$ back into data-like samples as $t:\infty \to 0$:
\begin{equation}
    \text{d}X_\tau = 
    \left[X_\tau  + 2 s\left(X_{\tau}, -\tau\right)\right] \text{d}\tau
    + \sqrt{2}\,\text{d}W_{\tau}, 
    \quad X_{-\infty} \sim \mu_\infty.
    \label{eq:reverse}
\end{equation}
Here, we define $\tau:=-t$, $\tau:-\infty \rightarrow 0$,
$\rho(x,t)$ denotes the forward density with initial distribution $\mu_0$,  
$s(\xi,t) := \nabla_\xi \log \rho(\xi,t) \in \mathbb{R}^D$ is the score function of the corrupted (forward) distribution given initial distribution $\mu_0$,
and $\text{d} W_s $ is again a multi-dimensional Wiener process. 
The central training objective of diffusion models is thus framed as \emph{learning the score function $s(x,t)$} \cite{songScoreBasedGenerativeModeling2021a}. In practice, a neural network $\mathbb{R}^D \times \mathbb{R} \to \mathbb{R}^D$ is used to approximate $s(x,t)$, which is then plugged into \eqref{eq:reverse} during sampling.

A key point is that the score function has a special mathematical structure: it is a conservative field. Neural networks used in practice are not constrained to produce conservative vector fields and, therefore, do not necessarily preserve this structure. This raises the central question of this study:  
\begin{quote}
Does a trained neural network actually learn a valid \emph{score function}, or merely a useful {vector field} for generative sampling?
\end{quote}

\subsection{Wasserstein Gradient Flow}

Wasserstein Gradient Flow (WGF) originates from the theory of optimal transport (OT), but it has become increasingly relevant for understanding modern generative models. Here, we  provide a brief overview and refer readers to the classic references \cite{ambrosioGradientFlowsMetric2008,figalliInvitationOptimalTransport2023} for comprehensive materials.

Recall the forward evolution of the probability density $\rho(x,t)$ under the FPE~\eqref{eq:FP}. In their seminal work, Jordan, Kinderlehrer, and Otto observed that an implicit Euler discretization of the FPE can be reinterpreted as a variational problem: each timestep corresponds to minimizing a free energy functional that combines Shannon entropy with a Wasserstein-2 distance penalty \cite{jordanVariationalFormulationFokkerPlanck1998}. This insight, known as the \emph{JKO scheme}, shows that the FPE can be understood as a gradient flow of entropy in the space of probability measures.

Building on this idea, Otto introduced a formal Riemannian calculus on the space of probability distributions, demonstrating that the FPE defines a steepest descent in  Wasserstein geometry \cite{Otto31012001}. This framework—now widely known as Otto calculus—precisely formalizes the notion that probability densities evolve like particles sliding down an energy landscape, but within the geometry induced by optimal transport. In addition, Otto also introduced the generalized Liouville equation (GLE)\footnote{We distinguish GLE from the ``continuity equation'',  a term commonly used in the field of OT. We make this distinction because continuity equations in physics can describe arbitrary conserved quantities (mass, energy, etc.), but the GLE specifically governs normalized probability density functions.\label{fn:GLE}}\cite{gerlich_verallgemeinerte_1973}.
Taken together, the JKO scheme and Otto’s formulation provide the foundation for WGF, unifying PDE evolution, entropy maximization, and optimal transport. One powerful result of WGF theory is:

\begin{quote}
While the sample paths of the diffusion process that FPE describes are fundamentally \emph{stochastic}, the marginal distribution\footnote{$\rho(\cdot,t)$ is referred to as the marginal distribution because it is only the distribution of $X_t$ at time $t$. It is a marginal distribution of the the joint distribution specified the stochastic process, $\rho\left(x_{t_1}, \ldots x_{t_N}\right)$.} of the paths at a specific time, $\rho(\cdot,t)$, is identical to the marginal distribution of the trajectories driven by a deterministic WGF.
\end{quote}

To see this, let us consider setting the energy functional as the sum of a quadratic potential and the negative Shannon entropy
\begin{equation}
    E\left\{\rho\left(\cdot,t\right)\right\} :=  \int \frac{x^2}{2}  \rho \left(x, t\right) \text{d} x + \int \rho \left(x, t\right) \log \rho \left(x, t\right)\, \text{d} x.  \label{eq:energy}
\end{equation}
Here, the first term accounts for the drift/advection and the second for the diffusion in the FPE \eqref{eq:FP}. The idea is to identify the \emph{steepest descent} direction functions that decrease the energy the most in the space of probability density functions induced by a deterministic velocity field $v(x,t)$. Applying $\text{d}/\text{d}t$ to the energy functional:
\begin{equation}
    \frac{\text{d}}{\text{d} t}E\left\{\rho\left(\cdot,t\right)\right\} = \int \frac{\delta E\left\{\rho\left(\cdot,t\right)\right\}}{\delta \rho (x,t)} \frac{\partial \rho \left(x,t\right)}{\partial t} \text{d} x, \label{eq:dEdt}
\end{equation}
where the functional variation of $E$ with respect to the density function $\rho$ can be explicitly computed:
\begin{align}
    \frac{\delta E\left\{\rho\left(\cdot,t\right)\right\}}{\delta \rho (x,t)} :={}&\frac{1}{\delta \rho (x,t)} \left[\int  \frac{x^2}{2} \delta  \rho\left(x,t\right) + \left(\rho +\delta \rho\right) \log \left(\rho +\delta \rho\right) \, \text{d} x - \int \rho \left(x, t\right) \log \rho \left(x, t\right)\, \text{d} x \right] \nonumber \\
    \sim {}& \frac{1}{\delta \rho (x,t)} \int \left[\frac{x^2}{2} +\log \rho(x,t) + 1\right]\delta  \rho\left(x,t\right)  \, \text{d} x =  \frac{x^2}{2}  + \log \rho(x,t). \label{eq:variation}
\end{align}
In the last two equations, we neglected higher-order $\mathcal{O}(\delta \rho(x,t))$ terms (using the asymptotic symbol $\sim$) and applied the normalization condition that the functional perturbation $\int \delta \rho(x,t)\,\text{d} x=0$ because $\int  \rho(x,t)\,\text{d} x = 1 =\int  \left(\rho+\delta \rho\right) (x,t)\,\text{d} x  $. 
Next, inserting GLE \cite{gerlich_verallgemeinerte_1973} (see footnote \ref{fn:GLE}):
\begin{equation}
    \partial_t \rho \left(x,t\right)= -\nabla_x \cdot \left[v(x,t) \rho(x,t) \right],
\end{equation}
and the functional variation \eqref{eq:variation} into \eqref{eq:dEdt} leads to
\begin{align}
    \frac{\text{d}}{\text{d} t}E\left\{\rho\left(\cdot,t\right)\right\} ={}& - \int \left[ \frac{x^2}{2} + \log \rho\left(x,t\right) \right] \nabla_x \cdot \left[v(x,t) \rho(x,t)  \right]\, \text{d} x \nonumber \\
    ={}& \int v(x,t)\cdot \left[ x + \nabla_x \log \rho\left(x,t\right) \right] \,  \rho(x,t) \, \text{d} x,
\end{align}
where we used integration by parts and assumed vanishing boundary terms. The above equation can be interpreted as an inner product of the functions $v(\cdot, t)$ and $\nabla_x \log \rho(\cdot,t) $ under the measure $\rho(\cdot,t)$. Clearly, the velocity field that corresponds to the steepest descent of the energy functional should align with the opposite direction of $\nabla_x \log(\cdot,t)$ (up to a global multiplicative constant):
\begin{equation}
    v_{\text{WGF}}(x,t) := - x - \nabla_x \log \rho (x,t) =  - x - s (x,t). \label{eq:v_WGF}
\end{equation}

The probability distribution of the resulting flow system with the above velocity field evolves under the GLE:
\begin{equation}
    \frac{\partial }{\partial t} \rho \left(x,t\right) = -\nabla_x \cdot \left[v_\text{WGF}(x,t) \rho(x,t) \right]  =\nabla_x \left[\left(x + \nabla_x \log \rho (x,t) \right) \rho\left(x,t\right)\right], \label{eq:target-GLE}
\end{equation}
which is exactly the FPE \eqref{eq:FP} describing the OU.

Song et al.~rediscovered the WGF velocity field \eqref{eq:v_WGF} through manipulating the FPE and noticing $\nabla_x \rho(x,t) = \rho(x,t)\, \nabla_x \log\rho(x,t)$. They used the term ``probability flow'', without referencing the JKO scheme, Otto calculus, and WGF. We believe it is beneficial to point out the origin of this theoretical framework, given its deeper connection to OT and the variational nature of the diffusion process. 

\section{Numerical experiments}

We now shift our focus to numerical experiments to verify the central question we have in score-based generative modeling: \emph{Are we learning the score function?} 

Due to the definition of the score function, $s(x,t):= \nabla_x \log \rho(x,t)$, the fundamental theorem of calculus (or generalized Stokes' theorem in high dimension) states that the line integral of the score function along a closed path in the state space has to be equal to zero:
\begin{equation}
\oint \vec{s}(x,t) \cdot \text{d}\vec{x} =  0. \label{eq:int}
\end{equation}
We will refer to \eqref{eq:int} as the \emph{integral constraint}. The second constraint, also following directly from the definition of the score function, states:
\begin{equation}
\frac{\partial}{\partial x_j} s_i (x, t) = \frac{\partial}{\partial x_i} s_j (x, t), \quad \text{for any pair }(i,j) \in \left\{1\ldots D\right\}^2. \label{eq:diff}
\end{equation}
We refer to \eqref{eq:diff} as the \emph{differential constraint}. Our goal is to numerically investigate wether either of the constraints are met in trained diffusion models.

\subsection{Models and datasets}

To present a minimal working example, we trained a MNIST diffusion model using a lightweight U‑Net implementation. The model is composed of ShuffleNet‑style residual bottlenecks and depthwise convolutions. The time indices are embedded, passed through an MLP, and added to the feature maps in each block. It employs simple encoder–decoder blocks with downsampling and upsampling and skip connections, keeping the model lightweight (around 4 MB). The implementation can be found at \cite{MNISTDiffusion}. We used the cosine schedule \cite{nicholImprovedDenoisingDiffusion2021} and a total discrete time index $T=1000$, which corresponds to observing time-homogeneous OU process \eqref{eq:OUprocess} at discrete times \cite{santos_understanding_2023}
\begin{equation}
    t_k =  -\frac{1}{2} \log \frac{f(k)}{f(0)}, \ f(k):=\cos \left( \frac{k/T+0.008}{1+0.008} \frac{\pi}{2}\right) 
\end{equation}

We also performed the same test with latent diffusion,  using a VAE with an $8\times 8$ latent space (implementation based on \cite{pytorch-vae}). The diffusion process employs the same network as before but acts in the latent space of the VAE.

The purpose of this experiment is to enable a comprehensive analysis with tractable computation, especially for evaluating the differential constraints. The results are presented in the following sections. We also observed a similar behavior for the CIFAR-10 dataset (Appendix \ref{appendix.CIFAR}).

\subsection{Integral constraints}
\begin{figure}[!t]
  \centering
\centering
    \includegraphics[width=\linewidth]{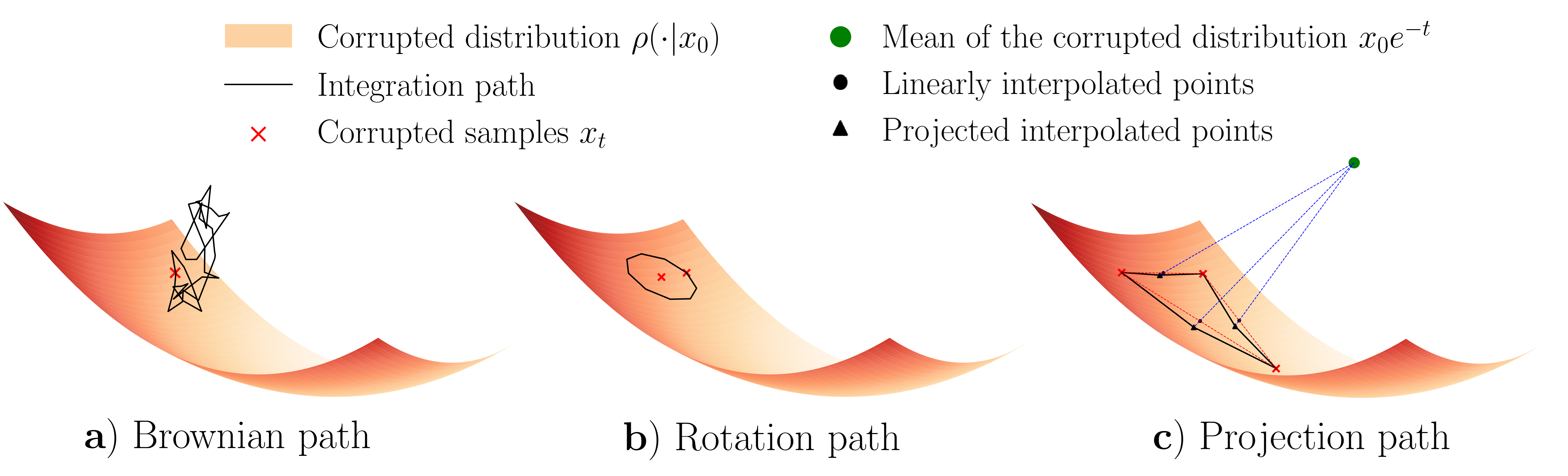}
\caption{ Mechanisms for assessing integral constraints. Illustration of the three mechanisms we used to construct closed paths for evaluating integral constraints within the high-density regions of the data distribution.}
% \YT{Remove the ``clean data point'' and projection line from (a) and (b)}}

  %\caption{ Three different mechanisms to generate closed-path for checking integral constraints in the high-volume region of the data distribution. \YT{Make the figures larger, currently there are too many white spaces. Fonts in the legends should be larger. For Brownian path, see if we can make the segment length scale about 1/4/ smaller (showing more ``wiggling''.) I like the rotation \& projection very much.}}
  \label{fig:path}
\end{figure}
To numerically check the integral constraint \eqref{eq:int}, we introduce three different mechanisms for generating closed paths on which the integral is evaluated:
\begin{itemize}[leftmargin=*]
    \item \textbf{Brownian path.} Starting from a corrupted sample $x_{t} \in \mathbb{R}^{D}$ generated by the forward diffusion, we perform a random walk on $\mathbb{R}^{D}$ using a Brownian bridge, which generates a path in $\mathbb{R}^{D}$ starting and ending at ${x}_{t}$. 
    The path of Brownian bridge is $X^\text{BB}_u = W_u  - u W_U / U$ with a fictitious time $u \in [0,U)$. We choose $U=9$, uniformly sample 1,000 discrete time steps in between, and add the resulting path to a forward sample $x_t$, i.e., $y_{u;t} = x_t + X^\text{BB}_u $.  This method does not guarantee that the path stays close to the the typical region induced by the forward process, as illustrated in Fig.~\ref{fig:path} (a). We include this path as a way to study  the behavior of out-of-distribution samples.
    \item \textbf{Rotation path.} Following the typical application of image corruption process, the corrupted sample $x_t={x}_0 e^{-t} + \sqrt{1-e^{-2 t}} \varepsilon $, where $\varepsilon \sim \mathcal{N}\left(0,I\right)$. We randomly pair each of the $D$ components of $\varepsilon$, so $(\varepsilon_i,\varepsilon_j)$ forms a two-dimensional vector. Then, we rotate each of the $D/2$ two-dimensional vectors with respect to the origin, i.e., $\varepsilon_i'(u)=\cos(2\pi u) \varepsilon_i + \sin(2\pi u) \varepsilon_j$ and $\varepsilon_j'(2\pi u)=- \sin(2\pi u) \varepsilon_i + \cos(2\pi u) \varepsilon_j$. Note that we rotate all $D/2$ pairs with the same ``angular velocity''. The resulting vector is used to generate a closed loop in the $x$-space, i.e., $y_{u;t} = {x}_0 e^{-t} + \sqrt{1-e^{-2 t}} \varepsilon'(u)$, $u:0\rightarrow 1$. With this construct, the probability density of noise realization $\varepsilon'(u)$ is identical to that of the original noise realization $\varepsilon$, ensuring the closed path in the $x$-space sits in the region where most of the probability mass is. 
    \item \textbf{Projection path.} We first generate multiple corrupted samples $x_{t}$ from the same initial $x_{0}$, then find a way to connect these points such that the connections lie in the typical set of corrupted distribution. In order to achieve this, we propose a simple mechanism: to connect two corrupted samples $x_t$ and $x'_t$, we first generate points that linearly interpolate between the two samples, and then project the interpolated points back to the corrupted distribution. Since Gaussian diffusion in high-dimensional space induces the structure of a thin shell around the clean samples, the projection can be carried out by projecting the samples radially back to the shell in $\mathbb{R}^D$, 
    whose radius is estimated either through Monte Carlo sampling (which we also know would be $\approx \sqrt{D}$ from asymptotic analysis). An illustrative schematic diagram is provided in Fig.~\ref{fig:path} (c).
\end{itemize}

\begin{figure}[!t]
  \centering
    \includegraphics[width=\linewidth]{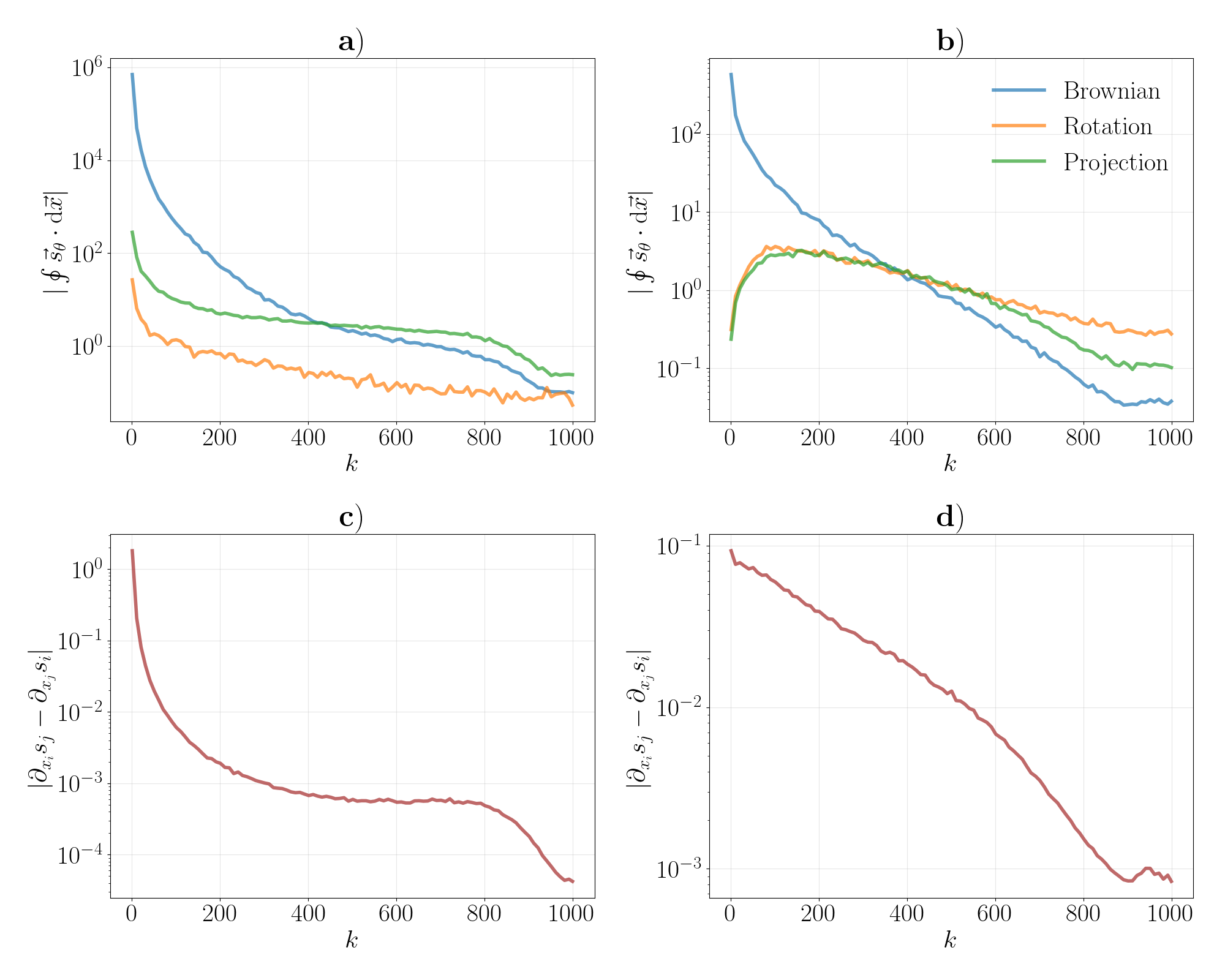}%
    
  \caption{Results of integral and differential constraints, as functions of discrete time index $k$: \textbf{a)} shows the absolute value of the integral condition $\oint \vec{s}_{\theta}\cdot {\rm d}\vec{x}$; \textbf{b)} presents the same quantity but for the latent dynamics; \textbf{c)} reports the differential condition $| \partial_{x_{i}}s_{j} -\partial_{x_{j}}s_{i}|$ in normal diffusion; \textbf{d)} shows the corresponding differential condition in latent diffusion.}
  % \YT{(1) use \texttt{plt.title('(a)')}, \texttt{plt.title('(b)')}\ldots and explain what they are in the caption here. (2) Fonts needs to be bigger. (3) Consolidate the font family---currently the labels, axes, numbers are all serif family (like Times New Roman) but the legends are sans-serif. (4) $x$-label should be $k$. (5) Change the $y$-label of (c,d) to $\left \vert\partial_i (s_\theta)_j - \partial_j (s_\theta)_i\right\vert$ (6) remove all the scaling $\sigma_t^2$.}}
  \label{fig:con}
\end{figure}

Figures \ref{fig:con} (a) and (b) show the results of evaluating the integral constraint using these three methods of generating closed paths.
%Since $2 s(x,t)$ is the actual drift in the reverse path, where the factor $2$ is the variance of the diffusion process and is a design choice, our main numerical results is multiplied by this factor to present the drift predicted by the neural network. 
Summary statistics of these distributions are provided in Fig.~\ref{fig:appendix.statistics} in the Appendix.

Clearly, the integral condition is not satisfied in the trained neural network. One may argue whether the magnitude matters to the reverse-time dynamics. To answer this, we notice that the score-induced drift $2 s(x,t) $ is added to a linear term $x(t)$ in \eqref{eq:reverse}; this provides us a non-dimensional quantity:
\begin{equation}
    \frac{ 2 \oint \vec{s}(\vec{y},t) \cdot \text{d}\vec{y}}{ \oint \left \vert \vec{y} \right\vert \left \vert \text{d} \vec{y} \right \vert} ,  \label{eq:non-dimensional}
\end{equation}
where $\vec{y}$ is a dummy vector looping over the generated path. Results of this quantity are presented in Figs. \ref{fig:appendix.mnist.con.int} and \ref{fig:appendix.statistics.norm} in Appendix, showing a significant deviation from 0. 

\subsection{Differential constraints}

Due to the intensive resources required to compute the full Jacobian matrix, we instead randomly sample $64$ components of the predicted score $s (x, t)$ and $64$ components of the corrupted samples $x_t$ to compute a $64\times 64$ sub-Jacobian matrix. The statistics were collected from 256 samples for each time step, and are presented in Fig.~\ref{fig:con} (c) and (d), both showing non-zero contributions.

\section{Discussion}

The numerical evidence clearly suggests that \emph{the trained neural network {\bf does not} learn the score function}, which is a conservative field. However, the trained network can definitely perform the generative task. The observation raises an interesting question: what is the trained neural network actually learning in order to perform the generative task? 

We here propose a bold hypothesis, leveraging the WGF theory, to understand what happens in the ``score-matching'' generative modeling. Our assertion is:
\begin{quote}
Existing diffusion modeling is better understood as modeling a normalizing flow \cite{chen_neural_2019}, through performing flow matching \cite{lipman_flow_2022} to the WGF velocity \eqref{eq:v_WGF}, rather than learning the reverse stochastic differential equation established by \cite{anderson_reverse-time_1982}, popularized by \cite{songScoreBasedGenerativeModeling2021a}.
\end{quote}

Contrary to typical flow-based models \cite{chen_neural_2019} which learn the velocity field by maximizing the end-to-end likelihood, 
the flow-matching method \cite{lipman_flow_2022} matches the neural velocity field to a target velocity field. The target velocity field is often analytically derived for a prescribed transport from the data distribution to an easy-to-sample distribution (often isotropic Gaussian distribution in high dimension), and evaluated on sampled training data. Here, we use the WGF induced by the energy functional \eqref{eq:energy} as the prescribed transport, and match the velocity field \eqref{eq:v_WGF}. More precisely, we only match the flow induced by the entropic term in \eqref{eq:energy}.

There are several advantages to understand the diffusion model as the flow-matching WGF. First, the ``probability flow'' is naturally included in the WGF framework. Secondly, we can formally bypass the necessity to invoke the reverse-time It\^o process, which can be confusing and counterintuitive---as will be seen below, within the WGF and Otto calculus framework, the deterministic probability flow ODE arises naturally, bypassing the need to explicitly route through Anderson’s reverse-time SDE. Finally, flow-matching WGF naturally explains why the trained neural flow, which fails to obey the differential and integral score conditions, can still perform in generative modeling. 

To see this, let us illustrate a self-consistent narrative of a flow-matching problem:
\begin{enumerate}[leftmargin=*]
    \item {\bf Optimization objective.} Our goal is to learn \eqref{eq:v_WGF} through flow-matching. We choose to minimize the $L^2$ error between the neural velocity and the entropy-induced velocity field in \eqref{eq:v_WGF}
    \begin{equation}
    \min_{\theta } \mathbb{E}_{k\sim \text{Unif}\left(\left\{1,2\ldots T\right\}\right)} \mathbb{E}_{x\sim \rho\left(\cdot, t\right)}\left \Vert v_\theta(\cdot,t_k) - s \left(\cdot,t_k\right) \right \Vert_2 
    \end{equation}
    \item {\bf Data generation.} Samples to perform Monte Carlo approximation of the above $L^2$-norm will be drawn from the distribution at time $t$, induced by the energy function \eqref{eq:energy}. Instead of using the WGF in the forward dynamics, which involves estimating $\log \rho$ in high dimension, we use the equivalent OU process \eqref{eq:OUprocess} to generate sample and more importantly, to compute \emph{analytically exact} $s(x,t)$ for matching the neural velocity field.
    \item {\bf Sampling/Inference.} To perform the generative task, terminal samples drawn from the isotropic Gaussian are transported from $t\rightarrow \infty$ to $t=0$ by integrating the ordinary differential equation backward in time. That is,  $\text{d}x(\tau)/{\text{d}\tau} = - v_\text{WGF}(x(\tau))= x(\tau) + \text{NN}(x(\tau),-\tau)$, where $\tau\equiv-t$, so signs flip relative to forward time. $x(\infty) \sim \mathcal{N}\left(0,I\right)$ and $\tau:-\infty \rightarrow 0$. The corresponding GLE \cite{gerlich_verallgemeinerte_1973} is
    \begin{equation}
    \frac{\partial }{\partial \tau} \rho \left(x,t\right) = - \nabla_x \left[\left(x + v_{\theta^\ast} \left(x,-\tau \right)\right) \rho\left(x,t\right) \right], \label{eq:trained-GLE}
    \end{equation}
    where $\theta^\ast $ stands for the trained neural weights.
\end{enumerate}
Operationally, the above descriptions are identical to applying the ``score-matching'' for training and performing ``probability flow'' for inference \cite{songScoreBasedGenerativeModeling2021a}. However, because of the deterministic nature of the WGF, we would not need to invoke the reverse-time stochastic process \cite{anderson_reverse-time_1982}. The simplicity is the first benefit of recognizing the existing approach as a Wasserstein Gradient Flow-Matching problem. 

By framing the learning as a flow-matching problem, it is most natural to weight each time equally, which is the \emph{de facto} training procedure for both discrete-time \cite{hoDenoisingDiffusionProbabilistic2020} and continuous-time \cite{songScoreBasedGenerativeModeling2021a} diffusion models. The procedure would seem \emph{ad hoc} if one aims to parameterize a neural network for learning the reverse-time diffusion process by a more theoretically grounded log-likelihood (more precisely, the bound of which) maximization as shown in \cite{sohl-dicksteinDeepUnsupervisedLearning2015}. As DDPM \cite{hoDenoisingDiffusionProbabilistic2020} pointed out, the log-likelihood approach involved weights which are not uniform in time; by removing such non-uniform weights, DDPM achieved a better performance by effectively solving a flow-matching problem. 

Next, assuming that we learn the WGF perfectly, we can treat the reverse-time WGF as a dynamical system:
\begin{equation}
\frac{\text{d} }{\text{d}\tau} x(\tau) = x(\tau) + \text{NN}(x(\tau), -\tau)=x(\tau) + \nabla_x \log (x(\tau), -\tau). 
\end{equation}
This system is identical to a Wasserstein Gradient Flow with the energy functional,
\begin{align}
    {}& E\left\{\rho\left(\cdot,\tau\right)\right\} =- \int  \frac{x^2}{2} \rho \left(x, \tau\right) \text{d} x - \int \rho \left(x, s\right) \log \rho \left(x, \tau\right)\, \text{d} x  \nonumber \\
    {}& =\underbrace{- \int  \frac{x^2}{2} \rho \left(x, \tau\right) \text{d} x - 2 \int \rho \left(x, \tau\right) \log \rho \left(x, \tau\right)\, \text{d} x}_{\text{Reverse-time drift}}  +  \underbrace{\int \rho \left(x, \tau\right) \log \rho \left(x, \tau\right)\, \text{d} x}_{\text{Reverse-time diffusion}}
    \label{eq:r_energy}
\end{align}
which is equivalent to the reverse-time It\^o process \eqref{eq:reverse}. This suggests that we would not need to invoke Anderson's seminal proof of the existence of the reverse diffusion \cite{anderson_reverse-time_1982} for generative task. This justifies the second advantage of the WGF framework. We remark, however, that to rigorously establish the equivalence of the forward and reverse \emph{path measures}, Anderson’s theory remains necessary. Nevertheless, because generative diffusion models only require consistency at the level of marginal densities, it is not necessary to invoke path measures in practice. We emphasize that our results concern density transport (marginals). We do not make claims about sample-path equivalence, which requires Anderson’s reverse-time construction. However, the corresponding reverse-time It\^o process not only can be used as a stochastic process for sampling, but also coincidentally the reverse-time process established by Anderson \cite{anderson_reverse-time_1982}. 

\begin{figure}[!t]
  \centering
    \includegraphics[width=\linewidth]{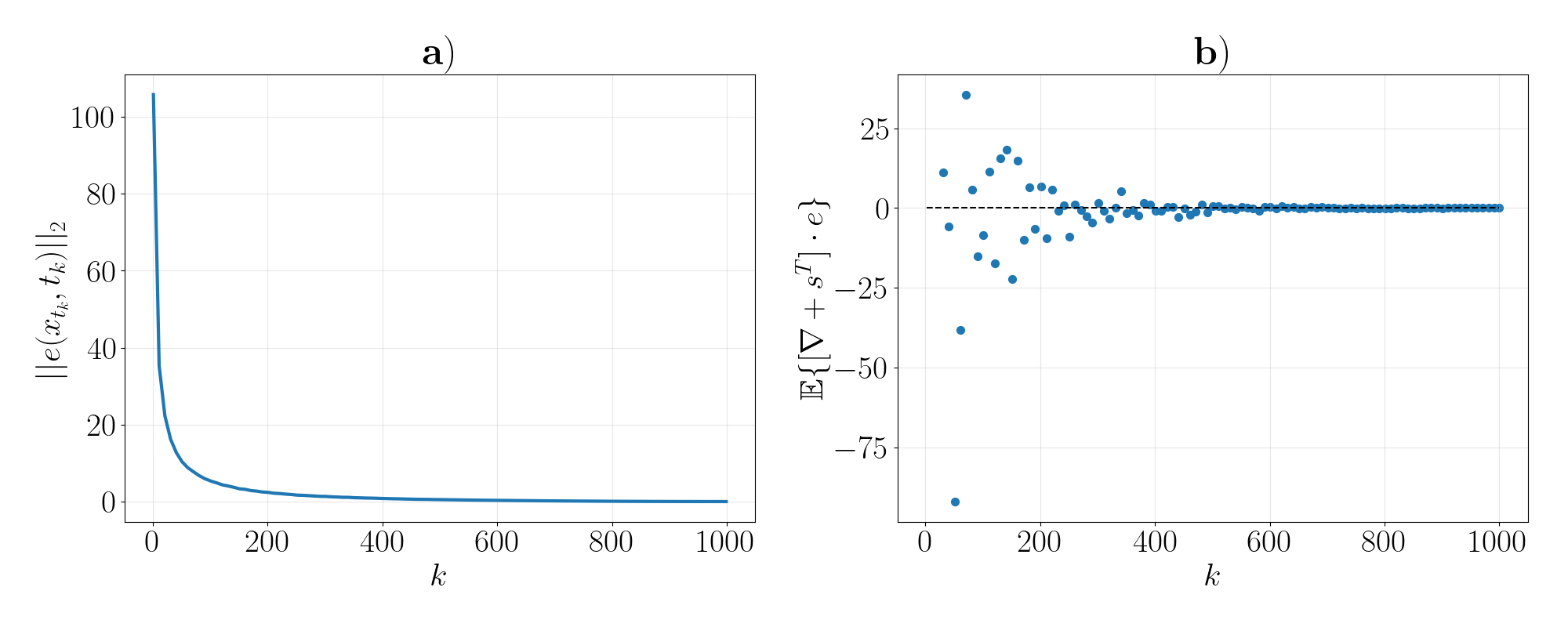}%
    
  \caption{\textbf{a)} L2 norm of $e(x,t)$ and \textbf{b)} Stein operator value of $e(x,t)$.}
  \label{fig:stein}
\end{figure}

Finally, as suggested by our numerical analysis, the neural network is \emph{not} learning a gradient of a scalar potential, i.e.~$\text{NN}(x,t)\ne s(x,t)$ for all $t$, both globally (because it violates the integral conditions) or locally (because it violates the differential conditions.) It is thus puzzling and challenging to analyze how the violations affect the reverse-time diffusion, and consequently the quality of the generated samples. The flow representation can bring some insight here. Suppose we use the trained, yet imperfect neural velocity field $\text{NN}(x,t) \approx \nabla_x \log \rho(x,t) $. Denote the error by $e(x,t):=s(x,t) - \text{NN}(x,t)$. Then, the GLE governing the distribution driven by the neural velocity field is
\begin{align}
    \frac{\partial}{\partial \tau} \rho(x,\tau) ={}& -\frac{\partial}{\partial x} \left[ \left( x + \text{NN}\left(x,-\tau\right) \right) \rho \left(x,-\tau\right) \right] \nonumber \\
    ={}&  -\frac{\partial}{\partial x} \left[ \left(x + s\left(x,-\tau\right)\right) \rho \left(x,-\tau\right) \right] + \frac{\partial}{\partial x} \left[ e(x,-\tau)  \rho\left(x,-\tau\right) \right] \nonumber \\
    ={}& -\frac{\partial}{\partial x} \left[ \left(x + s\left(x,-\tau\right)\right) \rho \left(x,-\tau\right) \right] \nonumber \\
    {}& + \left[\nabla_x \cdot e(x,-\tau) + s^T \left(x,-\tau\right)\cdot e \left(x,-\tau\right) \right] \rho(x,-\tau).
\end{align}
Immediately, we can identify a condition that if the error field $e(x,t)$ satisfies
\begin{equation}
    0 = \nabla_x \cdot e(x,t) + s^T \left(x,t\right) \cdot e\left(x,t\right),
\end{equation}
the induced distribution is identical to the true distribution. In other words, if $e(x,t)$ lives in the null kernel of the operator $\nabla_x + s^T(x,t) $, the trained neural network can perfectly perform the generative task, even if it is not perfectly capturing the score function. We remark that this vector operator is related to the Stein operator \cite{NIPS2016_b3ba8f1b} and is the key construct in several recent papers on sampling \cite{chenProjectedSteinVariational2020,fanPathGuidedParticlebasedSampling2024,tianLiouvilleFlowImportance2024}. In Fig. \ref{fig:stein}, we computed the error field on a trained latent diffusion model using forward generated samples, showing that indeed a significant $e(x,t)$ is induced (which is of order $10^2$, significant compared to the order $10^0$ of deterministic decaying flow, $\dot{x}(t)=-x(t)$), but the error field is statistically confined\footnote{We averaged over 256 randomly generated forward samples $x_t$. For each sample, the sufficient condition does not seem to be met but the average seems to agree, noting the significant variance for small $k$.} in the null kernel. This analysis suggests that:
\begin{quote}
    Even when $\text{NN}(x,t)$ is not the score function $\nabla_x \log p(x,t)$, the trained neural network can still be effective to perform generative modeling.
\end{quote}

We remark that this analysis is only possible by recognizing the underlying flow structure. We humbly acknowledge that we are not the first to propose the equivalence between  diffusion and  flow models: Song et al.~\cite{songScoreBasedGenerativeModeling2021a} recognized the ``probability flow/ODE'' and even suggested density and likelihood estimation, and very recently, Gao et al.~pointed out the resemblance between diffusion and flow models \cite{gaoDiffusionModelsGaussian2025}. Nevertheless, to our best knowledge, there have been no studies connecting diffusion models and normalizing flow parametrized by flow matching through the elegant theory of WGF and Otto calculus. The existing theories neither connect the flow operator to the Stein operator \cite{NIPS2016_b3ba8f1b}. Furthermore, the identification of a unified description between the diffusion models and WGF could inspire new forward random sampling (``data generation'') for training and regularizing flow-based models. 

To conclude, we advocate for this theoretical framework because, first, it was developed over 20 years ago, and yet has been largely ignored in the machine learning literature, and second, the setup is self-consistent, simple, concise, and elegant. We dedicate this work to the pioneers of WGF theory---Jordan, Kinderlehrer, and Otto---whose foundational insights continue to shape and inspire cutting-edge machine learning research today.

\section{Acknowledgment}
This work was performed at Los Alamos National Laboratory (LANL), operated by Triad National Security, LLC, for the National Nuclear Security Administration of the U.S. Department of Energy (Contract No.~89233218CNA000001). ABF was supported by the Applied Machine Learning Summer School, sponsored by the Information Science \& Technology Institute at LANL. JES and YTL were supported by the project ``Diffusion Modeling with Physical Constraints for Scientific Data'' (20240074ER), funded by the Laboratory Directed Research \& Development (LDRD) Program. YTL also gratefully acknowledges partial support from the LDRD projects ``Renovating Monte Carlo Methods and Codebases with Generative Models'' (20250391ER) and ``Assured Artificial Intelligence: Efficient and Scalable Guardrail Certification for Science and Security'' (20250048DR).
\printbibliography

@inproceedings{tianLiouvilleFlowImportance2024,
  title = {Liouville {{Flow Importance Sampler}}},
  booktitle = {Forty-First {{International Conference}} on {{Machine Learning}}},
  author = {Tian, Yifeng and Panda, Nishant and Lin, Yen Ting},
  year = {2024},
  month = jun,
  copyright = {CC0 1.0 Universal Public Domain Dedication},
  langid = {english}
}

@inproceedings{fanPathGuidedParticlebasedSampling2024,
  title = {Path-{{Guided Particle-based Sampling}}},
  booktitle = {Proceedings of the 41st {{International Conference}} on {{Machine Learning}}},
  author = {Fan, Mingzhou and Zhou, Ruida and Tian, Chao and Qian, Xiaoning},
  year = {2024},
  month = jul,
  pages = {12916--12934},
  publisher = {PMLR},
  issn = {2640-3498},
  langid = {english}
}

@misc{chenProjectedSteinVariational2020,
  title = {Projected {{Stein Variational Gradient Descent}}},
  author = {Chen, Peng and Ghattas, Omar},
  year = {2020},
  month = jun,
  eprint = {2002.03469},
  primaryclass = {cs},
  publisher = {arXiv},
  archiveprefix = {arXiv}
}

@inproceedings{NIPS2016_b3ba8f1b,
  title = {Stein Variational Gradient Descent: A General Purpose Bayesian Inference Algorithm},
  booktitle = {Advances in Neural Information Processing Systems},
  author = {Liu, Qiang and Wang, Dilin},
  editor = {Lee, D. and Sugiyama, M. and Luxburg, U. and Guyon, I. and Garnett, R.},
  year = {2016},
  volume = {29},
  publisher = {Curran Associates, Inc.}
}

@inproceedings{gaoDiffusionModelsGaussian2025,
  title = {Diffusion {{Models}} and {{Gaussian Flow Matching}}: {{Two Sides}} of the {{Same Coin}}},
  shorttitle = {Diffusion {{Models}} and {{Gaussian Flow Matching}}},
  booktitle = {The {{Fourth Blogpost Track}} at {{ICLR}} 2025},
  author = {Gao, Ruiqi and Hoogeboom, Emiel and Heek, Jonathan and Bortoli, Valentin De and Murphy, Kevin Patrick and Salimans, Tim},
  year = {2025},
  month = feb,
  langid = {english}
}

@misc{MNISTDiffusion,
  author       = {bot66},
  title        = {{MNISTDiffusion: Implement a MNIST (also minimal) version of denoising diffusion probabilistic model from scratch}},
  howpublished = {\url{https://github.com/bot66/MNISTDiffusion}},
  year         = {2022},
}

@misc{SLD,
  author       = {Won Seong},
  title        = {{Simple Latent Diffusion Model}},
  howpublished = {\url{https://huggingface.co/spaces/JuyeopDang/KoFace-AI}},
  year         = {2024},
}

@misc{pytorch-vae,
  author       = {sksq96},
  title        = {{A CNN Variational Autoencoder in PyTorch}},
  howpublished = {\url{https://github.com/sksq96/pytorch-vae/blob/master/vae.py}},
  year         = {2018},
}

@article{Otto31012001,
  title = {The Geometry of Dissipative Evolution Equations: {{The}} Porous Medium Equation},
  author = {Otto, Felix},
  year = {2001},
  journal = {Communications in Partial Differential Equations},
  volume = {26},
  number = {1-2},
  pages = {101--174},
  publisher = {Taylor \& Francis},
  doi = {10.1081/PDE-100002243}
}

@article{jordanVariationalFormulationFokkerPlanck1998,
  title = {The {{Variational Formulation}} of the {{Fokker--Planck Equation}}},
  author = {Jordan, Richard and Kinderlehrer, David and Otto, Felix},
  year = {1998},
  month = jan,
  journal = {SIAM Journal on Mathematical Analysis},
  volume = {29},
  number = {1},
  pages = {1--17},
  publisher = {{Society for Industrial and Applied Mathematics}},
  issn = {0036-1410},
  doi = {10.1137/S0036141096303359}
}

@book{figalliInvitationOptimalTransport2023,
  title = {An Invitation to Optimal Transport, {{Wasserstein}} Distances, and Gradient Flows},
  author = {Figalli, Alessio and Glaudo, Federico},
  year = {2023},
  edition = {Second edition},
  publisher = {EMS Press},
  address = {Berlin, Germany},
  isbn = {978-3-98547-550-6},
  langid = {english}
}

@book{ambrosioGradientFlowsMetric2008,
  title = {Gradient Flows: In Metric Spaces and in the Space of Probability Measures},
  shorttitle = {Gradient Flows},
  author = {Ambrosio, Luigi and Gigli, Nicola and Savar{\'e}, Giuseppe},
  year = {2008},
  edition = {2nd ed},
  publisher = {Birkh{\"a}user},
  address = {Basel},
  isbn = {978-3-7643-8722-8},
  langid = {english}
}

@misc{nicholImprovedDenoisingDiffusion2021,
  title = {Improved {{Denoising Diffusion Probabilistic Models}}},
  author = {Nichol, Alex and Dhariwal, Prafulla},
  date = {2021-02-18},
  number = {arXiv:2102.09672},
  eprint = {2102.09672},
  eprinttype = {arxiv},
  primaryclass = {cs, stat},
  publisher = {{arXiv}},
  archiveprefix = {arXiv},
  langid = {english}
}

@misc{songScoreBasedGenerativeModeling2021a,
  title = {Score-{{Based Generative Modeling}} through {{Stochastic Differential Equations}}},
  author = {Song, Yang and Sohl-Dickstein, Jascha and Kingma, Diederik P. and Kumar, Abhishek and Ermon, Stefano and Poole, Ben},
  date = {2021-02-10},
  number = {arXiv:2011.13456},
  eprint = {2011.13456},
  eprinttype = {arxiv},
  primaryclass = {cs, stat},
  publisher = {{arXiv}},
  archiveprefix = {arXiv},
  langid = {english},
  note = {Comment: ICLR 2021 (Oral)}
}

@misc{hoDenoisingDiffusionProbabilistic2020,
  title = {Denoising {{Diffusion Probabilistic Models}}},
  author = {Ho, Jonathan and Jain, Ajay and Abbeel, Pieter},
  date = {2020-12-16},
  number = {arXiv:2006.11239},
  eprint = {2006.11239},
  eprinttype = {arxiv},
  primaryclass = {cs, stat},
  publisher = {{arXiv}},
  archiveprefix = {arXiv},
  langid = {english}
}

@misc{sohl-dicksteinDeepUnsupervisedLearning2015,
  title = {Deep {{Unsupervised Learning}} Using {{Nonequilibrium Thermodynamics}}},
  author = {Sohl-Dickstein, Jascha and Weiss, Eric A. and Maheswaranathan, Niru and Ganguli, Surya},
  date = {2015-11-18},
  number = {arXiv:1503.03585},
  eprint = {1503.03585},
  eprinttype = {arxiv},
  primaryclass = {cond-mat, q-bio, stat},
  publisher = {{arXiv}},
  archiveprefix = {arXiv},
  langid = {english}
}

@misc{santos_understanding_2023,
    title = {Understanding {Denoising} {Diffusion} {Probabilistic} {Models} and their {Noise} {Schedules} via the {Ornstein}--{Uhlenbeck} {Process}},

    author = {Santos, Javier E. and Lin, Yen Ting},
    month = oct,
    year = {2023},
  eprint = {2311.17673},
  primaryclass = {stat,cond-mat, cs,math-ph},
  archiveprefix = {arXiv},

}

@article{anderson_reverse-time_1982,
    title = {Reverse-time diffusion equation models},
    volume = {12},
    issn = {03044149},
    %url = {https://linkinghub.elsevier.com/retrieve/pii/0304414982900515},
    doi = {10.1016/0304-4149(82)90051-5},
    language = {en},
    number = {3},
    journal = {Stochastic Processes and their Applications},
    author = {Anderson, Brian D.O.},
    month = may,
    year = {1982},
    pages = {313--326},
}

@article{gerlich_verallgemeinerte_1973,
    title = {Die verallgemeinerte {Liouville}-{Gleichung}},
    volume = {69},
    issn = {0031-8914},
    doi = {10.1016/0031-8914(73)90083-9},
    language = {en},
    number = {2},
    journal = {Physica},
    author = {Gerlich, G.},
    month = nov,
    year = {1973},
    pages = {458--466},
}

@inproceedings{lipman_flow_2022,
    title = {Flow {Matching} for {Generative} {Modeling}},
    url = {https://openreview.net/forum?id=PqvMRDCJT9t},
    language = {en},
    author = {Lipman, Yaron and Chen, Ricky T. Q. and Ben-Hamu, Heli and Nickel, Maximilian and Le, Matthew},
    month = sep,
    year = {2022},
}

@misc{chen_neural_2019,
    title = {Neural {Ordinary} {Differential} {Equations}},
  eprint = {1806.07366},
  primaryclass = {cs, stat},
  archiveprefix = {arXiv},
    author = {Chen, Ricky T. Q. and Rubanova, Yulia and Bettencourt, Jesse and Duvenaud, David},
    month = dec,
    year = {2019},

}

\newpage
\section{Appendix}
We provide more statistics of the non-dimensionalized quantity ${|\oint\vec{s}_{\theta} {\rm d}\vec{x}|}/{\oint |\vec{x}_{t}|\cdot|{\rm d}\vec{x}|}$ \eqref{eq:non-dimensional}, as well as experiment results on the CIFAR-10 dataset.
\subsection{More numerical results on MNIST}\label{appendix.mnist}
\begin{figure}
    \centering
    \includegraphics[width=0.98\linewidth]{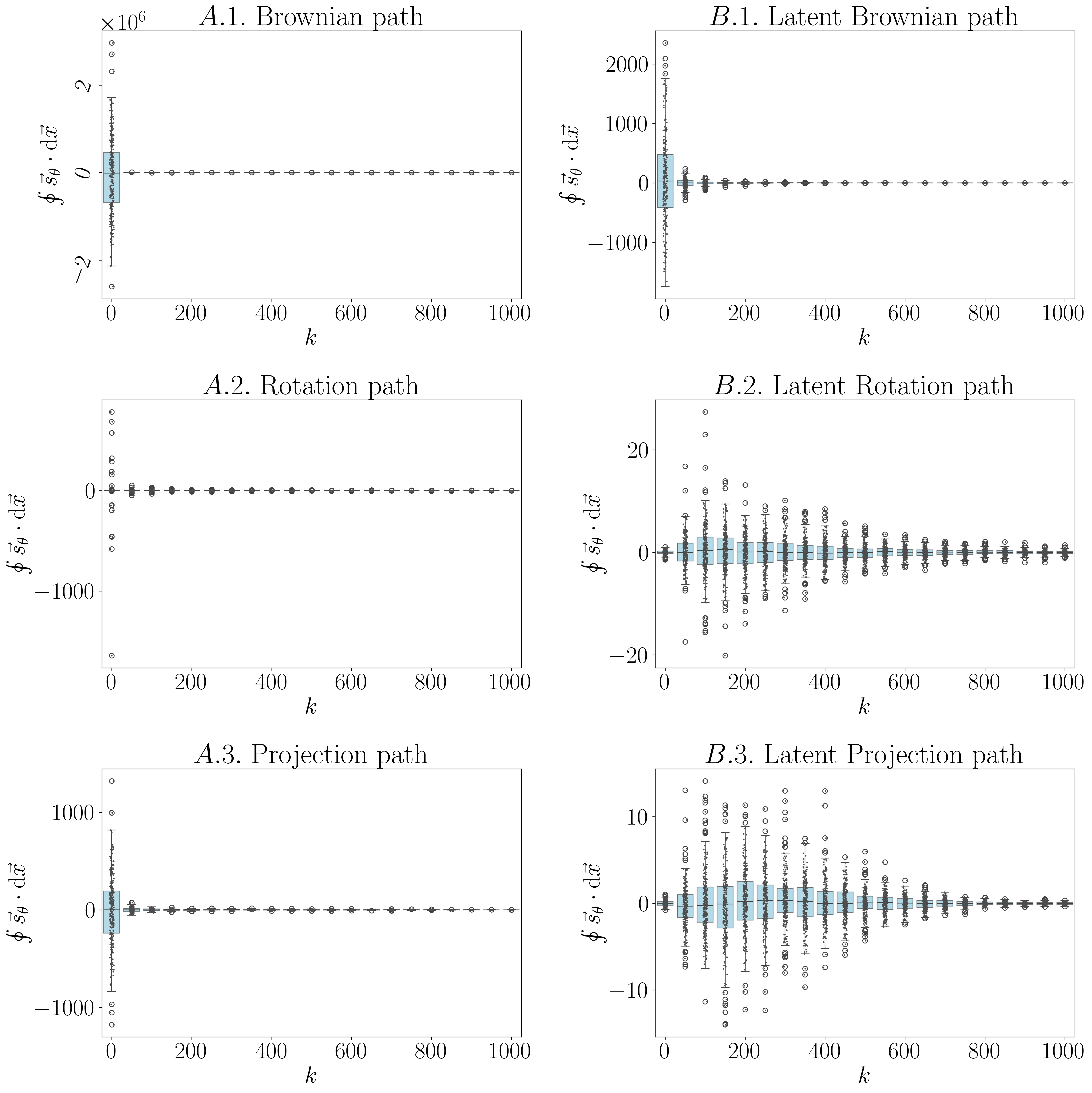}
    \caption{(MNIST) Summary statistics of $\oint \vec{s}_{\theta}\cdot {\rm d}\vec{x}$ calculated by different path-generating mechanisms, in normal and latent diffusions.}
    \label{fig:appendix.statistics}
\end{figure}

\begin{figure}
    \centering
    \includegraphics[width=0.98\linewidth]{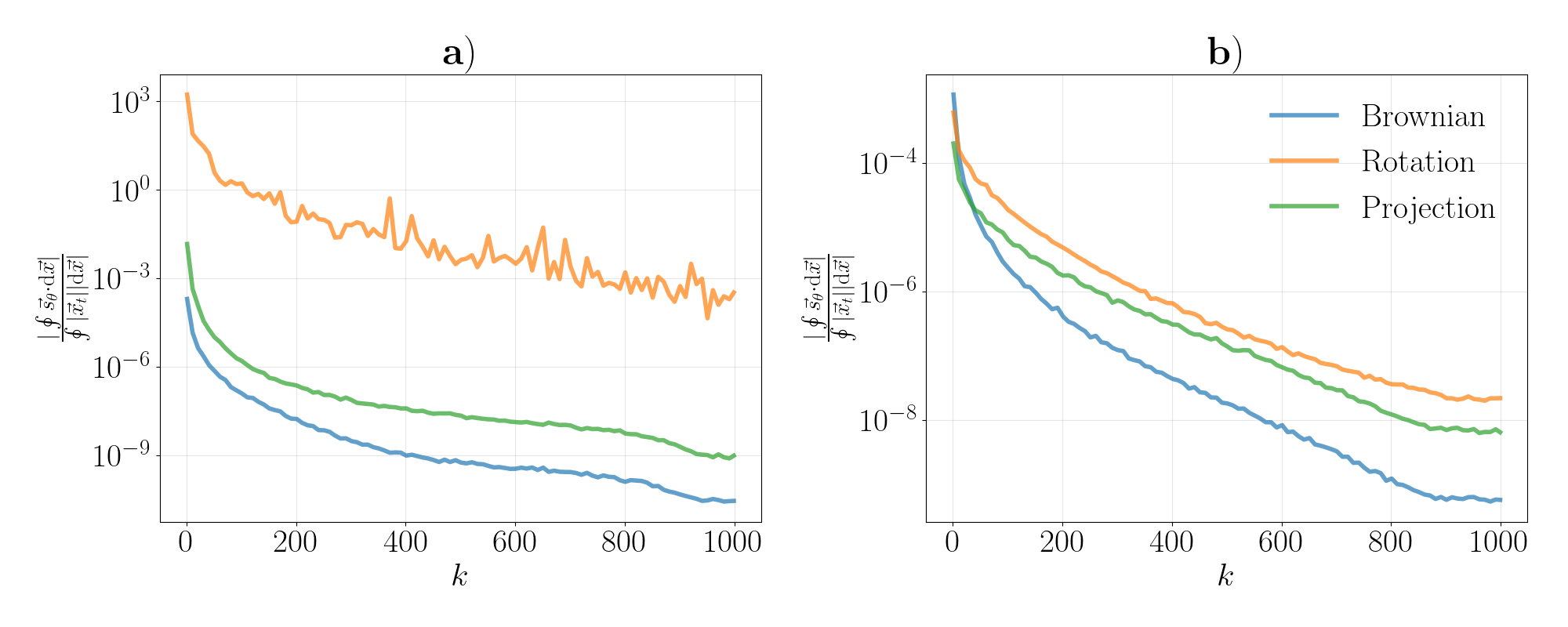}
    \caption{(MNIST) Results of integral constraints, as functions of discrete time index $k$: \textbf{a)} shows the absolute value of the integral condition $\oint \vec{s}_{\theta}\cdot {\rm d}\vec{x}$ normalized by the path length and the strength of the deterministic flow, $\oint |\vec{x}_{t}||{\rm d}\vec{x}|$; \textbf{b)} presents the same quantity but for the latent dynamics.}
    \label{fig:appendix.mnist.con.int}
\end{figure}

\begin{figure}
    \centering
    \includegraphics[width=0.98\linewidth]{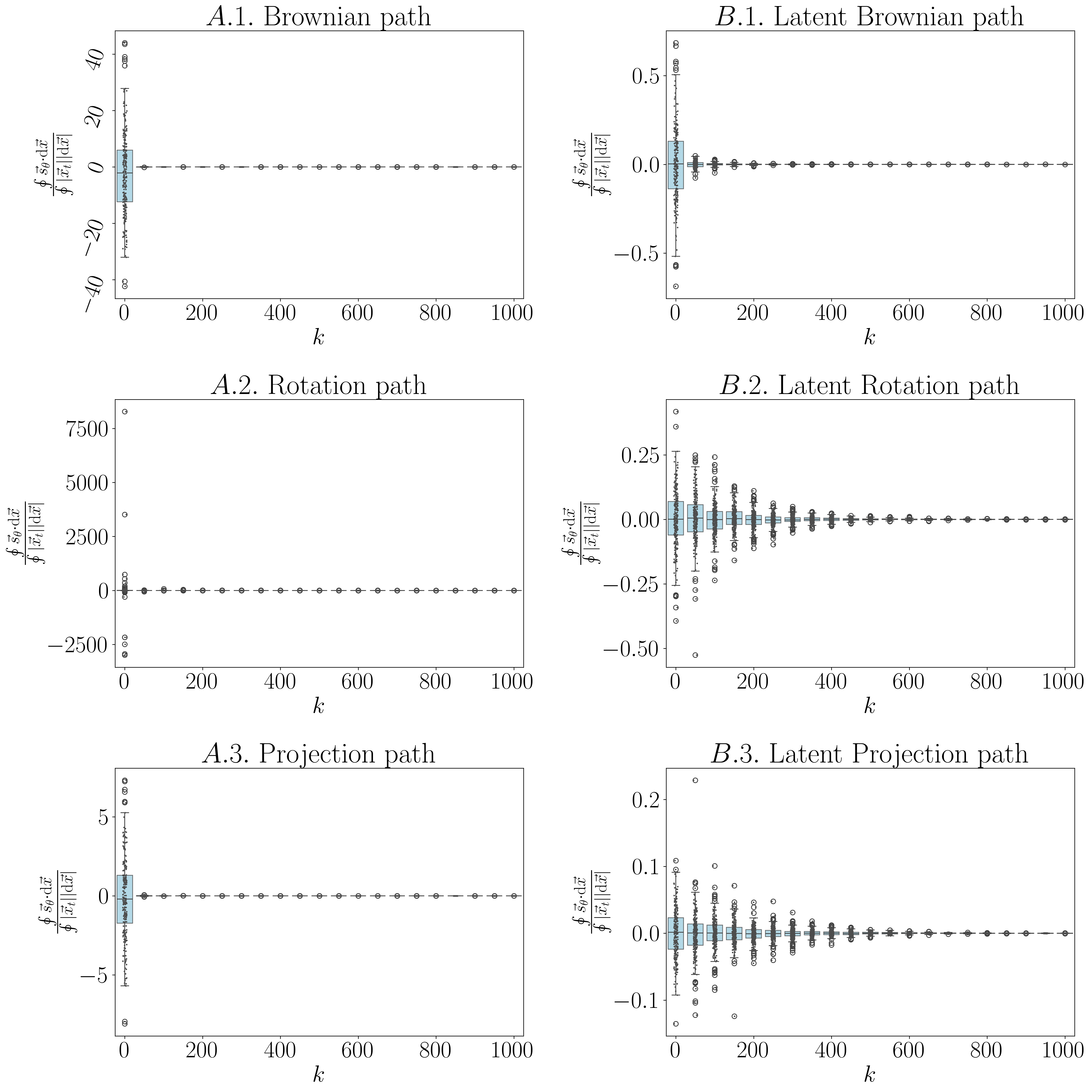}
    \caption{(MNIST) Summary statistics of ${|\oint\vec{s}_{\theta}\cdot {\rm d}\vec{x}|}/{\oint |\vec{x}_{t}||{\rm d}\vec{x}|}$ calculated by different path-generating mechanisms, in normal and latent diffusions.}
    \label{fig:appendix.statistics.norm}
\end{figure}

Refer to Figs. \ref{fig:appendix.statistics}, \ref{fig:appendix.mnist.con.int}, \ref{fig:appendix.statistics.norm}.

\subsection{Numerical results on CIFAR-10}\label{appendix.CIFAR}
\begin{figure}[!t]
  \centering
    \includegraphics[width=\linewidth]{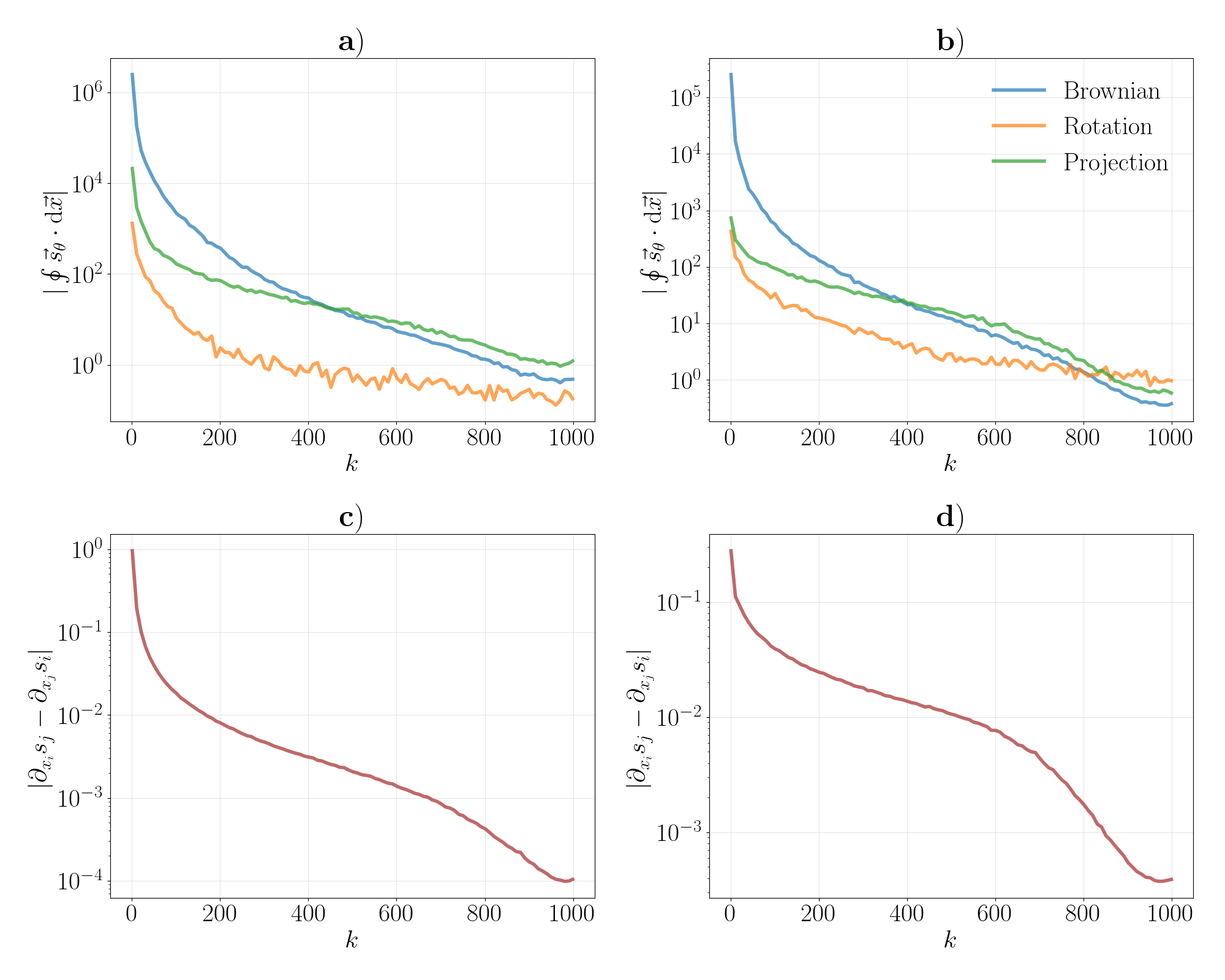}%
    
  \caption{(CIFAR-10) Results of integral and differential constraints, as functions of discrete time index $k$: \textbf{a)} shows the absolute value of the integral condition $\oint \vec{s}_{\theta}\cdot {\rm d}\vec{x}$; \textbf{b)} presents the same quantity but for the latent dynamics; \textbf{c)} reports the differential condition $| \partial_{x_{i}}s_{j} -\partial_{x_{j}}s_{i}|$ in normal diffusion; \textbf{d)} shows the corresponding differential condition in latent diffusion. }
  \label{fig:appendix.CIFAR1}
\end{figure}

\begin{figure}
    \centering
    \includegraphics[width=0.98\linewidth]{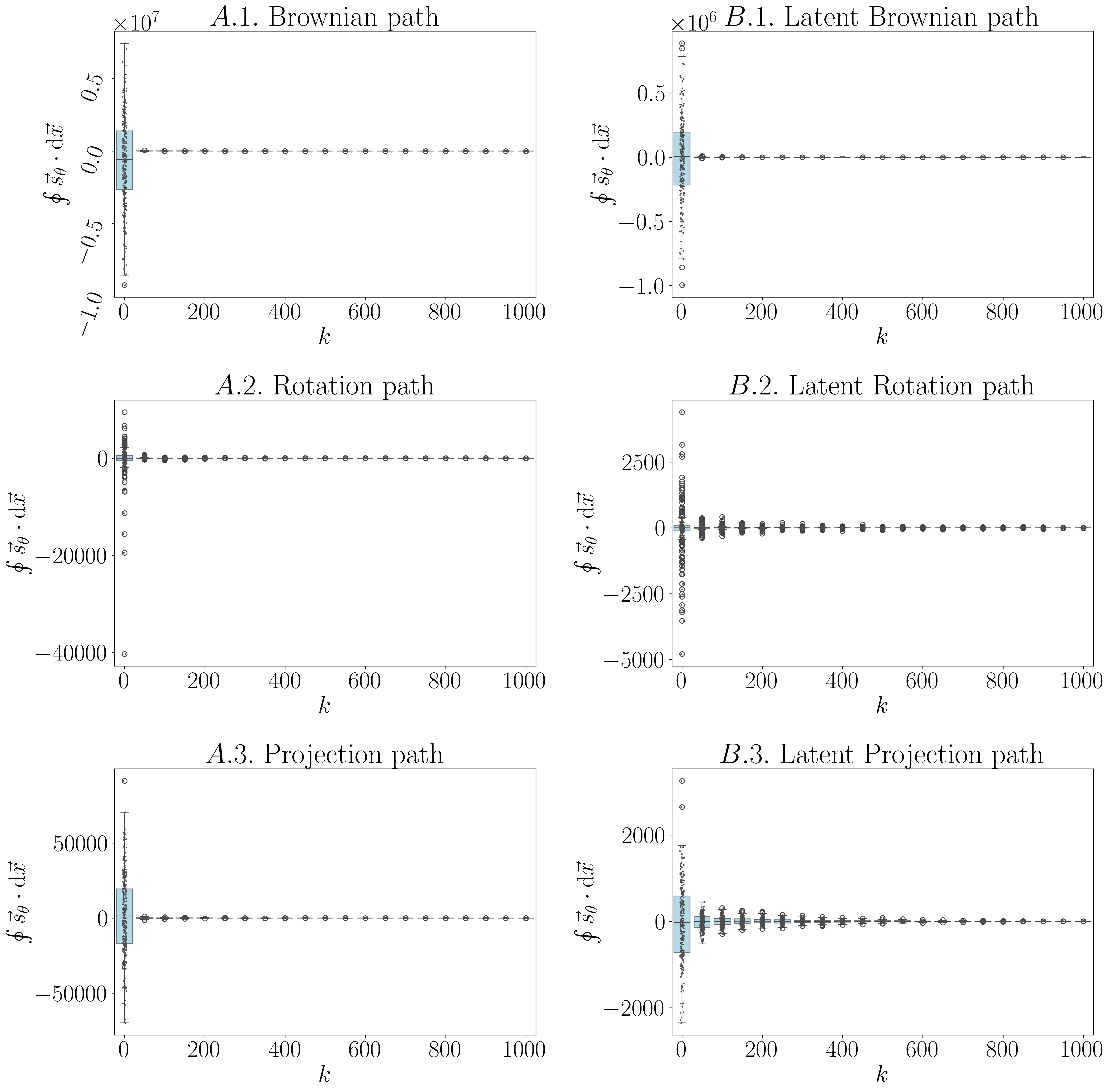}
    \caption{(CIFAR-10) Summary statistics of $\oint \vec{s}_{\theta}\cdot {\rm d}\vec{x}$ calculated by different path-generating mechanisms, in normal and latent diffusions.}
    \label{fig:appendix.CIFAR2}
\end{figure}

\begin{figure}
    \centering
    \includegraphics[width=0.98\linewidth]{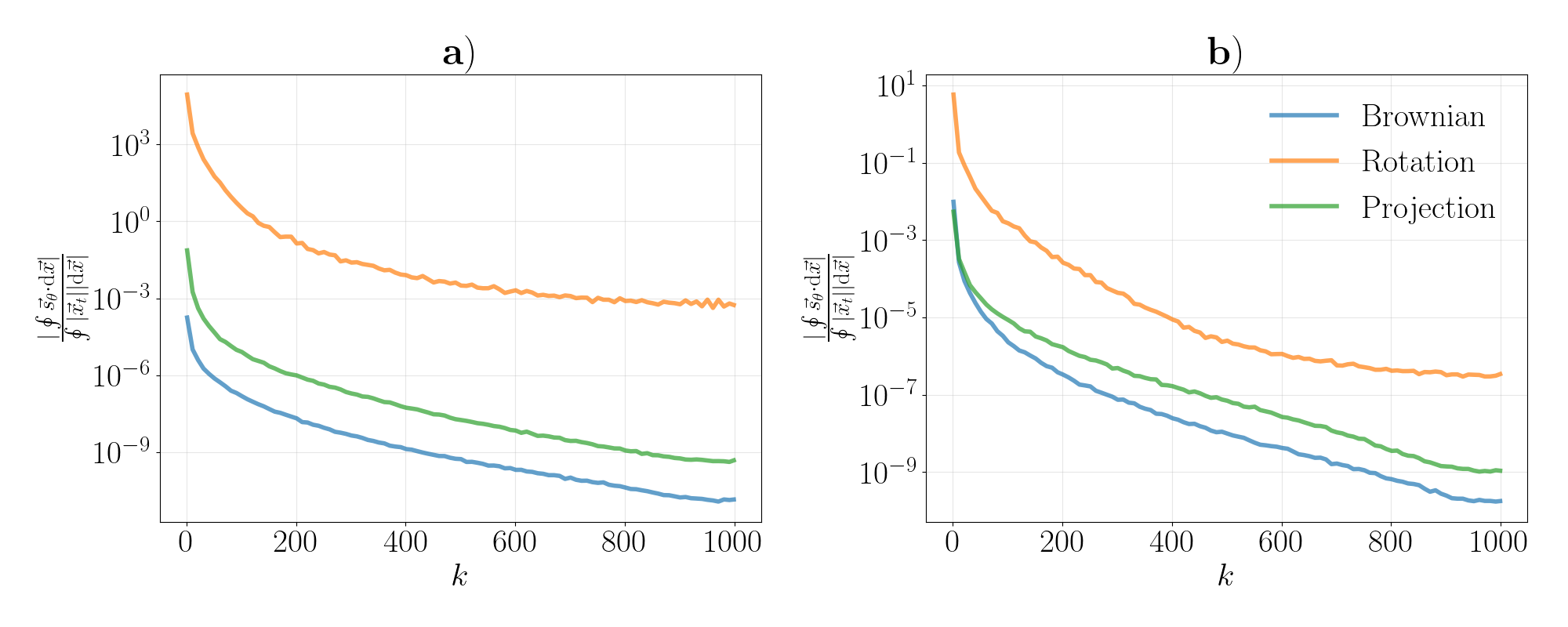}
    \caption{(CIFAR-10) Results of integral constraints, as functions of discrete time index $k$: \textbf{a)} shows the absolute value of the integral condition $\oint \vec{s}_{\theta}\cdot {\rm d}\vec{x}$ normalized by the path length and the strength of the deterministic flow,  $\oint |\vec{x}_{t}||{\rm d}\vec{x}|$; \textbf{b)} presents the same quantity but for the latent dynamics.}
    \label{fig:appendix.CIFAR3}
\end{figure}

\begin{figure}
    \centering
    \includegraphics[width=0.98\linewidth]{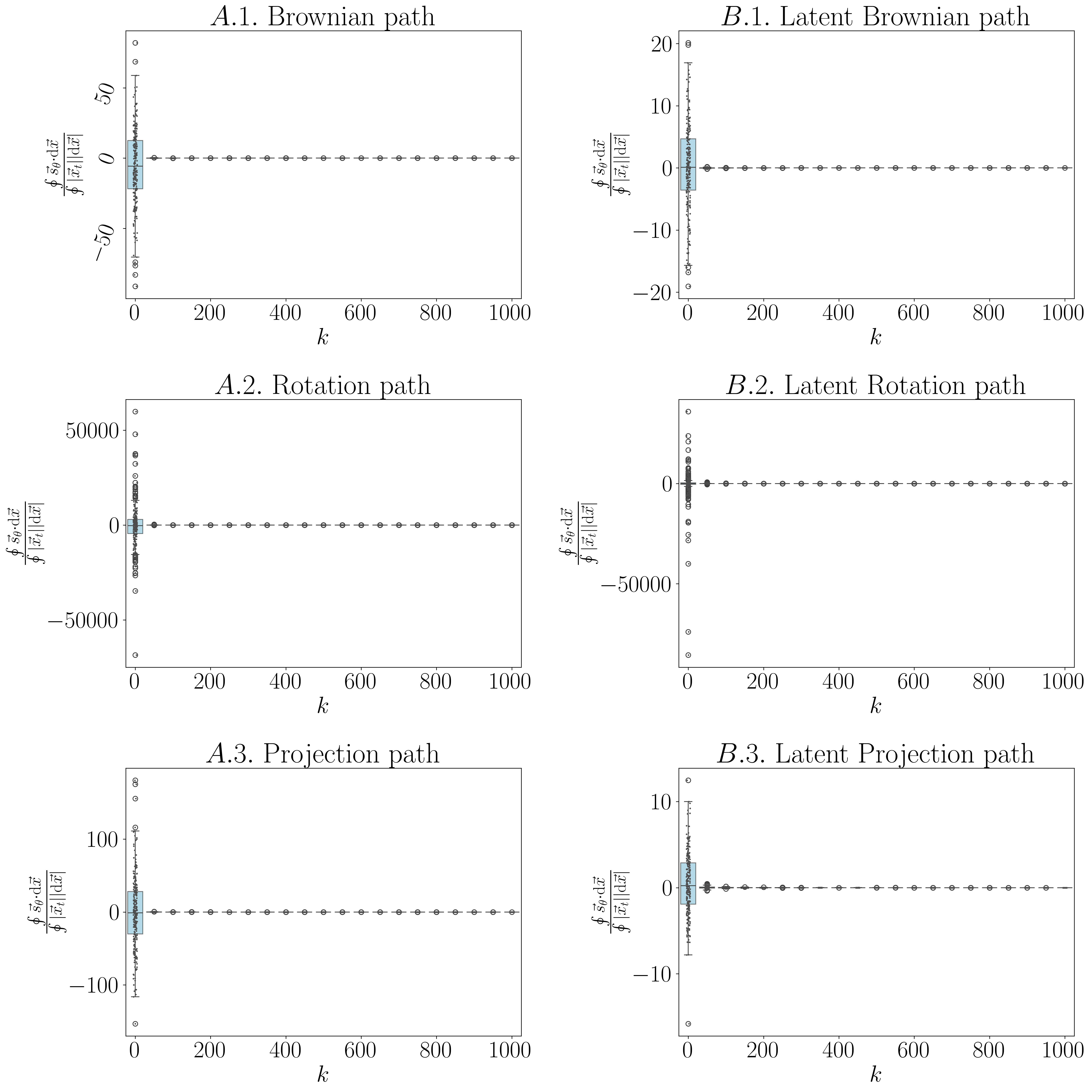}
    \caption{(CIFAR-10) Summary statistics of $|\oint\vec{s}_{\theta}\cdot {\rm d}\vec{x}|\oint |\vec{x}_{t}||{\rm d}\vec{x}|$ calculated by different path-generating mechanisms, in normal and latent diffusions.}
    \label{fig:appendix.CIFAR4}
\end{figure}

For CIFAR-10, we utilized the models from \cite{SLD}, it implements the standard DDPM and VAE with latent dimension of $3\times 16 \times 16$. We also tried training these models from scratch, which exhibits similar behaviors to the pretrained ones. Results are presented in Figs. \ref{fig:appendix.CIFAR1}, \ref{fig:appendix.CIFAR2}, \ref{fig:appendix.CIFAR3}, \ref{fig:appendix.CIFAR4}.

\end{document}